\documentclass[10pt,twocolumn,letterpaper]{article}

\usepackage{ijcb}
\usepackage{times}
\usepackage{epsfig}
\usepackage{graphicx}
\usepackage{amsmath}
\usepackage{amssymb}

\usepackage{algorithm}
\usepackage{algorithmic}
\usepackage{threeparttable}
\usepackage{float}  
\usepackage{subfigure}
\usepackage{url}

\ijcbfinalcopy 

\begin{document}

\title{Robust Face-Swap Detection Based on 3D Facial Shape Information}

\author{Weinan Guan$^{1,2}$, Wei Wang$^{2}$, Jing Dong$^{2}$, Bo Peng$^{2}$ \& Tieniu Tan$^{2}$ \\
	$^{1}$ School of Artificial Intelligence, University of Chinese Academy of Sciences\\
	$^{2}$ Center for Research on Intelligent Perception and Computing, CASIA\\
	\texttt{weinan.guan@cripac.ia.ac.cn,\{wwang,jdong,bo.peng,tnt\}@nlpr.ia.ac.cn}
}

\maketitle
\thispagestyle{empty}

\begin{abstract}
  Maliciously-manipulated images or videos - so-called deep fakes - especially face-swap images and videos have attracted more and more malicious attackers to discredit some key figures. Previous pixel-level artifacts based detection techniques 
  always focus on some unclear patterns but ignore some available semantic clues. Therefore, these approaches show weak interpretability and robustness. In this paper, we propose a biometric information based method 
  to fully exploit the appearance and shape feature for face-swap detection of key figures.
  The key aspect of our method is obtaining the inconsistency of 3D facial shape and facial appearance, and the inconsistency based clue offers natural interpretability for the proposed face-swap detection method.
  Experimental results show the superiority of our method in robustness on various laundering and cross-domain data, which validates the effectiveness of the proposed method.
\end{abstract}

\section{Introduction}
    Recently, with the developing of deep learning, especially Generative Adversarial Networks (GAN) \cite{NIPS2014_5423}, some studies of facial manipulation have shown rapid progress. The manipulated images or videos, so-called deep fakes, are always maliciously used to deceive the public. Face swapping, as one of deep fakes, is generated by replacing the target face with the source face but reserving the expressions of the target face. It attracts increasing attention of malicious people to slander some key figures.
    Therefore, deep-fake detection has arisen significant concerns.
    
    
    In the past years, deep-fake detection studies has seen a remarkable advance
    \cite{mirsky2020creation, tolosana2020deepfakes, 9065881}. Researchers exploit multiple clues to authenticate face-swap images and videos, such as details in the eye and teeth areas \cite{8638330}, the combination of local and global features \cite{8014963} and 3D head poses \cite{8683164}. For detecting deep-fake videos, the sequential information is also an important cue \cite{8639163, Sabir_2019_CVPR_Workshops}. Deep-fake detection is actually a statistical classification problem. Some classification neural networks can be applied to detect deep fakes and achieve satisfied performance. Previous work in deep-fake detection mostly captures the flaws of appearance as the detection clues in face-swap images and videos.
    
    In this paper, we propose to leverage the inconsistency between 3D facial shape and facial appearance information for protecting key figures from face-swap images and videos. 
    The facial appearance in a face-swap image or video is of the source individual, while 3D facial shape remains the target individual. The inherent flaw is employed to robustly detect face-swap images or videos, that 3D facial shape does not belong to the claimed person. Unlike previous work on modelling human behavior \cite{agarwal2020detecting, Agarwal_2019_CVPR_Workshops}, which requires the sequential information to follow head movements or trace facial expression, the biometric shape information can be extracted from a single image. Our proposed method shows the superior robustness against laundering counter-measures and on cross-domain data, compared to some previous pixel-level artifacts based approaches. And it can be generalized well in the wild. We summarize our main contributions as:
    \begin{itemize}
        \item We propose a novel clue and framework to detect the inconsistency between facial appearance and 3D facial shape information, and the inconsistency based clue provides the interpretability for the proposed face-swap detection method.
        \item We model the distance measurement of intra-class and inter-class data in order to fully exploit the data distribution for improving the detection performance.
        \item Extensive experiments against most laundering counter-measures and on cross-domain data demonstrate the superiority of our method in robustness and generalization.
    \end{itemize}
    
    \begin{figure*}
    	\begin{center}
    		\centerline{\includegraphics[width=0.8\textwidth]{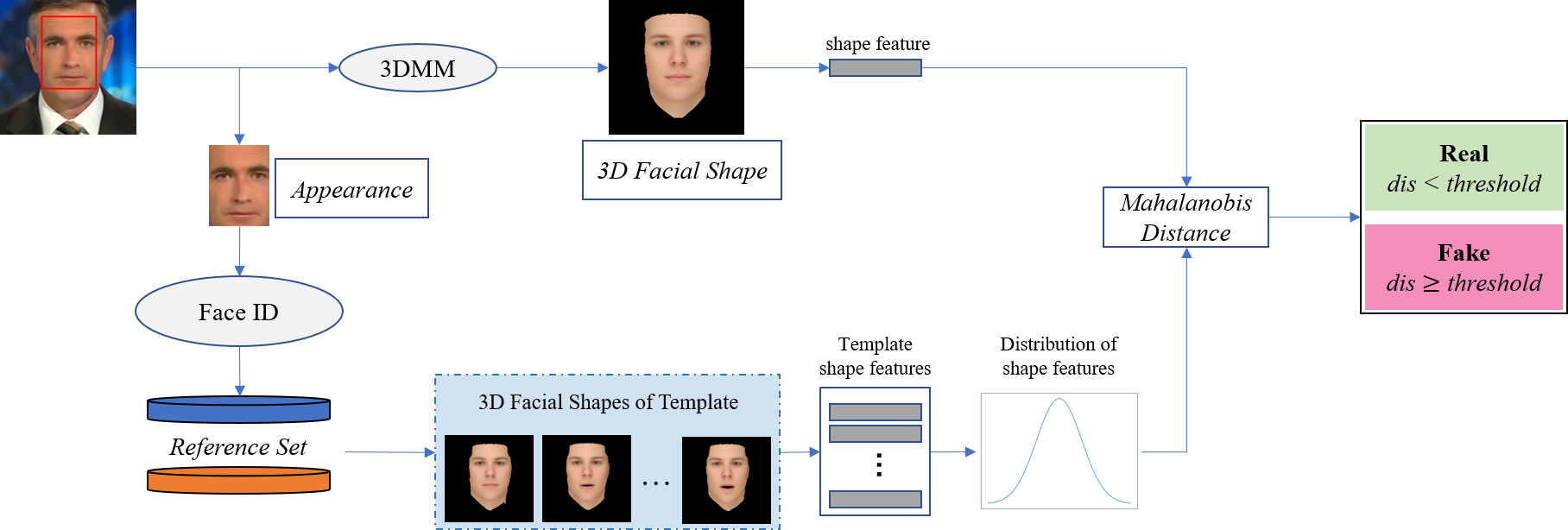}}
    	\end{center}
    	
    	\caption{An overview of our authentication pipeline.}
        \label{fig:architecture}
    \end{figure*}
\section{Related Work}
    In this section, some previous work related to face swapping is introduced. First, we describe some manipulation methods, and then discuss some detection methods.

\subsection{Face-swap Manipulation}
    Face swapping has attracted extensive attention of researchers. Several manipulation methods are proposed to generate compelling fake images and videos, including $\textit{FaceSwap}$ \cite{FaceSwap}, $\textit{Deepfakes}$ \cite{Deepfakes}, and $\textit{FSGAN}$ \cite{Nirkin_2019_ICCV}. 
    
    $\textit{FaceSwap}$ is a graphics-based approach to transfer the face region from a source video to a target video \cite{Rossler_2019_ICCV}. This method performs face swapping by fitting the 3D model to landmarks in the source face and aligning it to the target face. The rendering and color correction are then applied to improve the visualization of fake images and videos.
    
    Unlike traditional approaches that are based on computer-graphics, some methods rely on deep-learning algorithms, especially GANs, to create face-swap images and videos. For two specific individuals, $\textit{Deepfakes}$ trains two autoencoders with a shared encoder to reconstruct the source and the target faces, respectively \cite{Rossler_2019_ICCV}. The face-swap image is produced from the source face by the trained encoder and target decoder. The fake face is then blended with the target image using Poisson image editing \cite{10.1145/1201775.882269, Rossler_2019_ICCV}. However, this technique is needed to train a model for every pair of faces. 
    $\textit{FSGAN}$ 
    is subject agnostic, which can be applied to pairs of faces without training on those faces \cite{Nirkin_2019_ICCV}. In $\textit{FSGAN}$, face segmentation and reenactment are introduced to adjust for both pose and expression variations. Moreover, a face blending network is proposed to preserve the target skin color and lighting conditions. 
    
    With the advances of manipulation methods, several previous flaws in deep fakes have been concerned and fixed. Therefore, some previous clues are not efficient and the visualization of deep fakes is more realistic than before.
    
\subsection{Face-swap Detection}
    By the continuous development of technology for detecting face-swap images and videos, numerous novel methods are emerged to capture various flaws of face-swap deep fakes \cite{tolosana2020deepfakes}. Here, we concentrate on some current methods based on different clues, including pixel-level artifacts based low-level approaches and semantic clues based high-level approaches \cite{agarwal2020detecting}.
    
    Pixel-level artifacts based low-level approaches detect deep fakes relying on the marks of generation process, which is neglected by human eyes\cite{agarwal2020detecting}. Researchers utilize common classification neural networks to automatically capture the pixel-level artifacts, such as 
    $\textit{XceptionNet}$ \cite{Chollet_2017_CVPR}, and $\textit{EfficientNet}$ \cite{tan2019efficientnet}. 
    $\textit{XceptionNet}$ , as an modified structure inspired by $\textit{Inception}$ \cite{szegedy2015going}, has a good performance on $\textit{FaceForensics++}$ Dataset \cite{Rossler_2019_ICCV}. Furthermore, in the $\textit{Deepfake Detection Challenge (DFDC)}$,  $\textit{XceptionNet}$ is proposed as a baseline method \cite{dolhansky2019deepfake}. Compared to  $\textit{XceptionNet}$, $\textit{EfficientNet}$ has a similar or better performance of manipulation detection. Thus, $\textit{XceptionNet}$ and $\textit{EfficientNet}$ always tend to be the backbone models for modified methods in deep-fake detection. Nevertheless, pixel-level artifacts based methods always suffer from simple laundering counter-measures which can easily destroy the measured artifacts (e.g., additive noise, recompression, smoothing) \cite{Agarwal_2019_CVPR_Workshops}.  
    The performance of trained models is decreased dramatically on laundering data, and the models are even non-effective on cross-domain data.
     
    Semantic clues based high-level approaches utilize some semantically meaningful features for detecting manipulations\cite{agarwal2020detecting}, like 
    the color clues \cite{mccloskey2018detecting}, 3D head poses \cite{8683164}, and blending boundaries \cite{li2020face}. 
    In \cite{mccloskey2018detecting}, the authors prove the difference of color between the images from a real camera and a network. Researchers in \cite{8683164} observe the inconsistency between 3D head poses estimated from the facial landmarks and the central face region. In another attempt \cite{li2020face}, the blending boundaries are detected in deep fakes generated by a shared step in the process of face manipulation, which blends the altered face into an existing background image. The clue is proved effective in detecting deep fakes. However, most of the previous high-level semantic clues focus on the appearance of deep fakes that is also the concern of deep-fake generation. Therefore, for creating more realistic deep fakes, researchers tend to eliminate the appearance flaws. Then, new deep-fakes against these detection methods will be generated.
     
    In our method, we describe a robust clue for protecting key figures from face-swap deep fakes. The inconsistency between appearance and 3D facial shape is leveraged by the method as the detection clue against most laundering counter-measures and cross-domain data. Furthermore, as a result of the application of different modal clues, our detection method has the power to defend the attack of generation methods only focusing on appearance.
     
\section{The proposed method} \label{method}
    Fig~\ref{fig:architecture} shows the proposed detection framework. In order to capture the inconsistency of 3D facial shape and facial appearance in face-swap images and videos, we first utilize 3DMM (3D morphable model) \cite{10.1145/311535.311556} to extract 3D facial shape features of face-swap images and template videos (Subsection~\ref{sub:3DMM}). Then, we calculate $\textit{Mahalanobis Distance}$ between the shape features of suspected images and corresponding templates, and the distance is further utilized to authenticate the suspected images by comparing with the fixed threshold (Subsection~\ref{sub:measurement}).
    
\subsection{Face-Swap Artifacts} \label{sub:clue}
    In the process of confrontation between generation and detection methods, the facial appearance in face-swap images and videos has been constantly enhanced. For this reason, it is difficult for face-swap detection only relying on the facial flaws. However, face swapping only performs manipulation on the facial region. It focuses on changing the facial appearance in the target faces. Therefore, it ignores to transfer other biometric information from source faces to target faces. We concentrate on the inherent flaw in face-swap deep fakes, and
    build our framework for face-swap detection based on a cross-modal clue.
    
    Specifically, we utilize the inconsistency between facial appearance and 3D facial shape in face-swap deep fakes for detection. For a specific individual, his/her facial shape should not change significantly in diverse images and video footage. In face-swap deep fakes, current methods only replace the target faces with the source faces but retain facial shapes of target images. Hence, the flaw of the cross-modal inconsistency provides us a novel clue for face-swap detection.
    
\subsection{3D Facial Shape} \label{sub:3DMM}
    3D morphable model is proposed to estimate 3D facial model from a single face \cite{10.1145/311535.311556}. 
    In this work, we utilize the 3DMM fitting model \cite{7533097, Zhu_2015_CVPR} 
    to reconstruct the 3D face with facial shape and texture information. 
    It only fits the 3D facial shape based on the correspondences of 3D and 2D facial landmarks.
    
    Based on \cite{7533097} and \cite{Zhu_2015_CVPR}, the facial shape with expression can be estimated as follows:
    \begin{equation}
    S = \bar{S} + A_{id}\alpha_{id} + A_{exp}\alpha_{exp} \label{eq:3DMM_fitting}
    \end{equation}
    where $\bar{S}$ denotes the mean vector of 3D facial shape, $A_{id}$ and $A_{exp}$ are the matrices comprised by the principal components of 3D facial shape variances representing identification and expression respectively, and $\alpha_{id}$ and $\alpha_{exp}$ denote the weighted coefficients for $A_{id}$ and $A_{exp}$. Thus, $\alpha_{id}$ and $\alpha_{exp}$ are the determinants of various facial shape and expression, respectively. The method further projects 3D facial landmarks into 2D plane. Moreover, the correspondences in the original face as possible are achieved by the projected 2D facial landmarks. To this end, the projection process is described as:
    \begin{equation}
    s_{2D}(P, R, t, \alpha_{id}, \alpha_{exp}) = PR(s_{3D}(\alpha_{id},  \alpha_{exp}) + t) 
    \label{eq:3DMM_shape}
    \end{equation}
    where $R$ and $t$ denote the rotation matrix and the translation vector of 3D facial landmarks $s_{3D}$, respectively. Furthermore, $P$ is the orthographic projection matrix. We set $\theta = \left\{P,R,t,\alpha_{id}, \alpha_{exp} \right\}$ as the set of parameters to be optimized. The fitting process is to minimize the distance between the ground truth of 2D facial landmarks and the projected landmarks.
    
    In our framework, we only concentrate on the facial shape related to the identity irrespective of expression. Therefore,  $\alpha_{id}$ is the only required parameter. For simplicity, we also perform feature selection on $\alpha_{id}$ to obtain facial shape features. The details of feature selection are shown in Subsection~\ref{sub:feature_selection}.
    
\subsection{Inconsistency Measure} \label{sub:measurement}
    Here, we introduce a measurement method to calculate the distance between the shape features of a manipulated image and the corresponding template. Then, the distance is compared with a given threshold for detecting deep fakes. In our method, due to the inaccuracy of estimation, the template is enrolled with a set of 3D facial shape features and a particular distribution is modeled for the features. Therefore, for the comprehensive utilization of the template information, we use $\textit{Mahalanobis Distance}$ for measurement. Compared to other common measurements, $\textit{Mahalanobis Distance}$ is scale-invariant and uses the relations of various features. Moreover, it can compute the distance between a point and a distribution. Consequently, in our approach, we utilize $\textit{Mahalanobis Distance}$ to calculate the distance between 3D facial shape features in a manipulated image and the distribution modeled by the corresponding template. The formula of $\textit{Mahalanobis Distance}$ is described as follows:
    \begin{equation}
        D(\Vec{x}) = \sqrt{(\Vec{x}-\Vec{\mu})^T \Sigma^{-1} (\Vec{x}-\Vec{\mu})}
    \label{eq: mahalanobis_distance} 
    \end{equation}
    where, $\Vec{x}$ denotes the 3D facial shape feature vector of the manipulated image, and $\Vec{\mu}$ and $\Sigma$ are the mean vector and the covariance matrix of the corresponding template, respectively. The formula shows that $\textit{Mahalanobis Distance}$ utilizes the covariance matrix to integrate the relations of various features and distribution information of a template. It is noticed that regarding the inverse covariance matrix, the number of images in the template must not be less than the feature dimensions.
    
     
     
    After measuring the distance, a threshold needs to be determined in the training phase for detecting deep fakes. It is obvious that the 3D facial shape features of a genuine image should be close to the corresponding template. Thus, an image is classified as fake if the computed distance between it and its template is above the given threshold. We tune and fix our threshold with the criterion of approximate accuracy between genuine and manipulated images.

    \begin{figure} 
	\centering  
	\subfigure[]{
		\label{fig:C23}
		\includegraphics[width=0.07\textwidth]{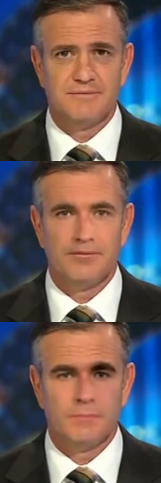}}
	\subfigure[]{
		\label{fig:C40}
		\includegraphics[width=0.07\textwidth]{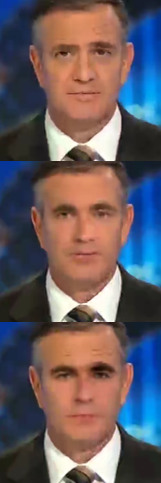}}
	\subfigure[]{
		\label{fig:GN001}
		\includegraphics[width=0.07\textwidth]{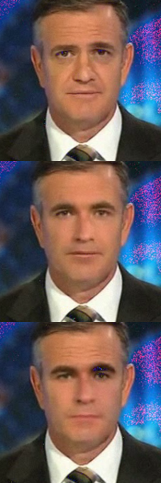}}
	\subfigure[]{
		\label{fig:GN01}
		\includegraphics[width=0.07\textwidth]{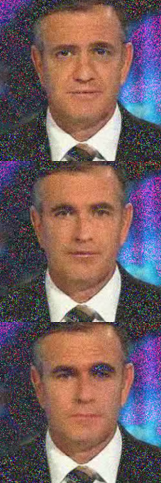}}
	\subfigure[]{
		\label{fig:GS07}
		\includegraphics[width=0.07\textwidth]{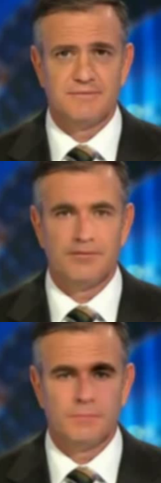}}
	\subfigure[]{
		\label{fig:GS13}
		\includegraphics[width=0.07\textwidth]{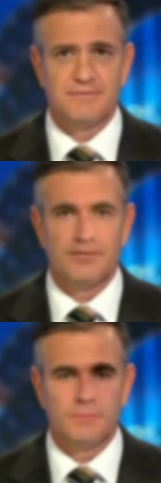}}
	\caption{Shown are sample images generated by various manipulation methods and attacked by different laundering counter-measures. The top row is the genuine images, the medium row denotes the fake images created by $\textit{FaceSwap}$, and the bottom column is the images generated by $\textit{Deepfakes}$.
	(a) $\textit{CRF23}$ (b) $\textit{CRF40}$ (c) $\textit{Gaussian Noise (var=0.001)}$ (d) $\textit{Gaussian Noise (var=0.01)}$ (e) $\textit{Gaussian Smoothing (std=1.4)}$ (f) $\textit{Gaussian Smoothing (std=2.3)}$}
	\label{fig:low_quality_1}
\end{figure}

\section{Experiments} \label{experiment}
\subsection{Dataset}

    Since we mainly concentrate on face-swap detection in this paper, deep fakes created by $\textit{FaceSwap}$ \footnote{\url{https://github.com/MarekKowalski/FaceSwap/}.} and $\mathit{Deepfakes}$ \footnote{\url{https://github.com/deepfakes/faceswap}.} from FaceForensics++ Dateset \cite{Rossler_2019_ICCV} are used as our experimental dataset. Fig~\ref{fig:C23} and Fig~\ref{fig:C40} show some examples of the dataset. In the following, more details about the dataset are provided.
    
    \noindent \textbf{FaceForensics++ Dataset~\cite{Rossler_2019_ICCV}.}	FaceForensics++ Dataset is a large-scale facial forgery dataset, which has $1,000$ pristine videos collected from the Internet. The face-swap videos in the dataset are generated by $\textit{FaceSwap}$ and $\mathit{Deepfakes}$, respectively. $\textit{FaceSwap}$ is a graphics-based method to swap the faces of a source video to a target video, while $\mathit{Deepfakes}$ is a deep learning based method of face swapping. In the generation process of face-swap videos, the $1,000$ pristine videos are randomly split into $500$ pairs for face swapping. In every pair, face swapping is conducted by $\textit{FaceSwap}$ and $\mathit{Deepfakes}$, respectively. Thus, both $\textit{FaceSwap}$ and $\mathit{Deepfakes}$ generate $1,000$ fake videos respectively.
    
    \noindent \textbf{Facial Shape Registration.} Since we try to utilize facial shape information to authenticate the subject featured by his/her facial appearance, the shape features of each protected person should be registered as a template.  We first calculate the shape coefficients $\alpha_{id}$s (Eq.~\ref{eq:3DMM_shape}) of each individual from the first five seconds of the corresponding video (the first seven seconds for a video with fps less than $20$). We further perform feature selection on $\alpha_{id}$s to obtain our facial shape features and then we determine the mean vector and the covariance matrix of the features for each person as his/her facial shape template.
    
    \noindent \textbf{Dataset Splitting.} For impartial comparison with previous methods mentioned in FaceForensics++ Dataset, we arrange our training, testing, and validation data with the official procedure. In the training phase, only genuine videos and face-swap videos manipulated by $\mathit{FaceSwap}$ are used for training and validating. Furthermore, they all undergo $\textit{FFmpeg}$ \footnote{FFmpeg. \url{http://ffmpeg.org/}} $\textit{CRF23}$ compression for saving disk space. Moreover, we further utilize $\mathit{Deepfakes}$ as cross-domain data for assessment in the testing phase. Fig~\ref{fig:low_quality_1} shows some examples of our data.
    
    \noindent \textbf{Laundering Attacks.} $\textit{CRF40}$ compressed videos are also provided by FaceForensics++ Dataset with lower quality compared to $\textit{CRF23}$ (see in Fig~\ref{fig:C40}). To assess the robustness of our method, we also attempt more laundering attacks, such as additive noise and smoothing. Gaussian smoothing ($\textit{GS}$) are used to $\textit{CRF23}$ videos by $\mathit{OpenCV}$ \footnote{OpenCV. \url{https://opencv.org/}} package. The sizes of the the gaussian kernels are $7$ and $13$ corresponding to standard deviations 1.4 and 2.3, respectively (Fig~\ref{fig:GS07} and Fig~\ref{fig:GS13}). Furthermore, gaussian noise ($\textit{GN}$) with zero means are added in each channel of RGB frames. We take into account two various noise levels with the variance of $0.001$, and $0.01$, respectively, as shown in Fig~\ref{fig:GN001} and Fig~\ref{fig:GN01}.
    
\subsection{Experiment Setup}

    Our proposed method conducts face-swap detection through the inconsistency between 3D facial shape and appearance. Specifically, a face-swap image always retains the facial shape of the target face but preserves the appearance of the source face. Hence, this clue is capable of face-swap detection as a result of the different facial shapes in different individuals. And we conduct the first experiment to validate the differences of facial shape in different individuals. As shown in the red part of Fig~\ref{fig:Mah}, we calculate $\textit{Mahalanobis Distance}$ between the mean vector of 3D facial shape features in a template and other templates. And the green part shows the discrepancy between an individual and his/her template. It is noticed that $\textit{Mahalanobis Distance}$ over 100 are reduced to 100 for simplicity. The figure separates the two part strongly supporting our hypothesis.

    Furthermore, we try to examine the resilience of our method to laundering attacks and cross-domain data and compare the results with some previous pixel-level artifacts based methods. In particular, we further set $\textit{XceptionNet}$ and $\textit{EfficientNet}$ as comparison methods, where $\textit{XceptionNet}$ shows the best performance in the FaceForensics++ Dataset. Specifically, we replace the output layer of $\textit{XceptionNet}$ and $\textit{EfficientNet}$ with a fully connected layer of single output and a $\textit{Sigmoid}$ activation is appended. In our approach, the direction of training is to obtain a threshold with approximate accuracy between genuine and manipulated images, which is used in the testing phase.

\begin{figure} 
	\centering  
	\subfigure{
		\includegraphics[width=0.225\textwidth]{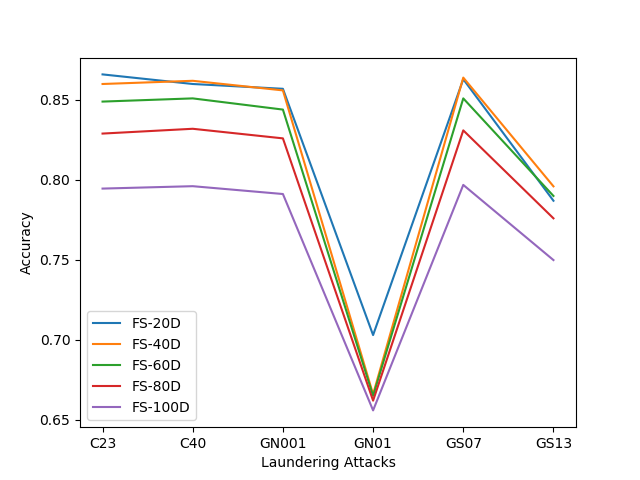}}
	\subfigure{
		\includegraphics[width=0.225\textwidth]{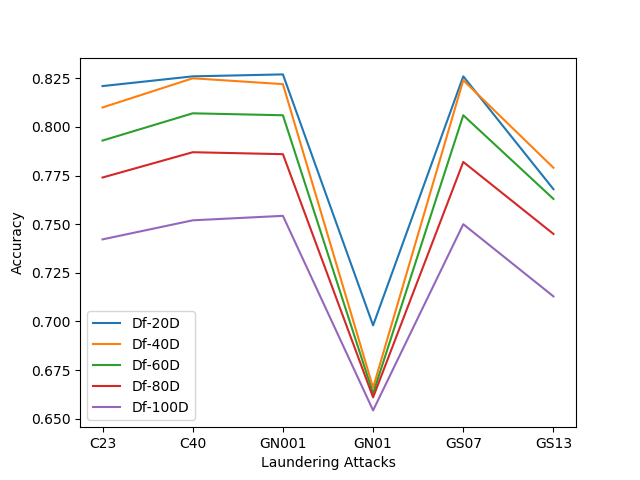}}

	\caption{The performance of five feature selection strategies on laundering (top) and cross-domain (bottom) testing data.}
	\label{fig:feat_selection}
\end{figure}

\subsection{Ablation Study}
    \textbf{Shape Feature Selection.} \label{sub:feature_selection} For simplicity, we perform feature selection based on 3D facial shape features. We set 100 frames as the minimum duration of each registered video. Then, due to the employment of $\textit{Mahalanobis Distance}$, the maximum dimension of the shape features is limited to 100.
    
    We design five feature selection strategies, selecting the top 20, 40, 60, 80, 100 dimensions of the $\alpha_{id}$s as facial shape features, respectively. They are assessed on the validation dataset and the strategy with best performance is employed in our method. According to Table~\ref{tab:feature_selection}, in which the values at every position denote the accuracy of all samples ($\textit{ACC}$), the feature
    selection strategy of taking the top 20 dimensions has the best performance for face-swap detection on validation data. Furthermore, we also assess the five strategies on all testing data. In Fig~\ref{fig:feat_selection}, our strategy is still superior for authenticating laundering and cross-domain data. It verifies the effectiveness of our feature selection strategy.
    
\begin{figure} 
	\centering  
	\subfigtopskip=2pt
	\subfigbottomskip=2pt
	\subfigure[$\textit{Cosine Distance}$]{
		\label{fig:cos}
		\includegraphics[width=0.20\textwidth]{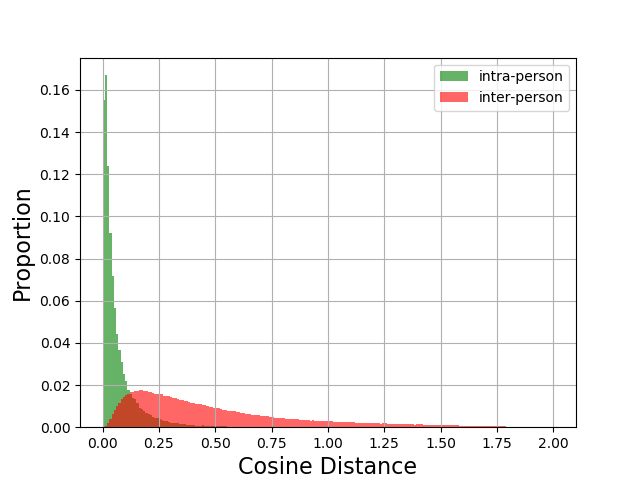}}
	\subfigure[$\textit{Mahalanobis Distance}$]{
		\label{fig:Mah}
		\includegraphics[width=0.20\textwidth]{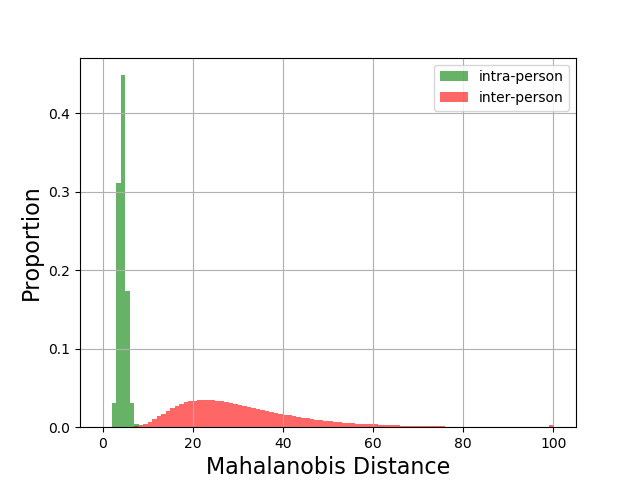}}

	\caption{$\textit{Cosine Distance}$ (a) and $\textit{Mahalanobis Distance}$ (b) between 3D facial shape features in templates are shown here. The red/green part corresponds to the distance between mean vectors/3D facial features of a template/a frame and corresponding/other templates.}
	\label{fig:cos_mah_hist_distance}
    \end{figure}

    \textbf{Distance Metrics.} We also report the results of another measurement method $\textit{Cosine Distance}$. According to Fig~\ref{fig:cos_mah_hist_distance}, we use various inconsistency measurement methods to examine the distinguishability between various characters under the same experimental setting.
    
    Specifically, in Fig~\ref{fig:cos}, for the green part, we determine $\textit{Cosine Distance}$ between facial shape features of all frames in the registered videos and the mean vectors of the corresponding templates. The red part is $\textit{Cosine Distance}$ from the mean vectors of other templates. In Fig~\ref{fig:Mah}, the measurement method is substituted with $\textit{Mahalanobis Distance}$.
    
    In Fig~\ref{fig:cos}, the red and green parts have a large overlapping area indicating that in our task, the facial shape features of different individuals cannot be distinguished by $\textit{Consine Distance}$  and the given threshold. This problem is significantly weakened with $\textit{Mahalanobis Distance}$.
    
    We further conduct comparison experiments on testing data, including laundering data and cross-domain data. The experimental results are shown in Table~\ref{tab:laundering} and Table~\ref{tab:cross-domain}. Compared to our method with $\textit{Consine Distance}$ (Ours(C)), our method with $\textit{Mahalanobis Distance}$ (Ours(M)) outperforms on most testing data. The results of the two experiments demonstrate that $\textit{Mahalanobis Distance}$ is more appropriate for our task than other measurement methods.
    
    \begin{table}
        \centering
		\begin{tabular}{c c c c c c}
			\hline
			 $\textit{Dimension}$ & $\textit{20-D}$ & $\textit{40-D}$ & $\textit{60-D}$ & $\textit{80-D}$ & $\textit{100-D}$\\
			\hline 
			$\mathbf{\textit{ACC}}$ & $\mathbf{0.873}$
             & 0.866 & 0.854 & 0.834 & 0.805 \\

			\hline
		\end{tabular}
    \caption{Accuracy of different feature selection on validation data.}
    \label{tab:feature_selection}
    \end{table}
    
    \begin{table}\footnotesize
        \centering
		\begin{tabular}{c c c c c c}
			\hline
			& $\textit{XceptionNet}$ & $\textit{EfficientNet}$ & \textbf{Ours(M)} & \textbf{Ours(C)}\\
			\hline 
			$\mathbf{*\textit{CRF23}}$ & 0.974 & $\mathbf{0.980}$ & 0.866 & 0.750 \\
			$\textit{CRF40}$ & 0.855 & 0.816 & $\mathbf{0.860}$ & 0.754\\
			$\textit{GN(var=0.001)}$ & 0.790 & 0.842 & $\mathbf{0.857}$ & 0.746\\
			$\textit{GN(var=0.01)}$ & 0.600 & 0.596 & $\mathbf{0.703}$ & $\mathbf{0.706}$\\
			$\textit{GS(std=1.4)}$ & 0.770 & 0.720 & $\mathbf{0.863}$ & 0.751\\
			$\textit{GS(std=2.3)}$ & 0.694 & 0.666 & $\mathbf{0.787}$ & 0.720\\

			\hline
		\end{tabular}
    \caption{\footnotesize{Comparison of the Laundering Data. $\textit{ACC}$}}
    \label{tab:laundering}
    \end{table}
    
    \begin{table}\footnotesize
        \centering
		\begin{tabular}{ccccc}
			\hline
			& $\textit{XceptionNet}$ & $\textit{EfficientNet}$ & \textbf{Ours(M)} & \textbf{Ours(C)}\\
			\hline 
			$\textit{CRF23}$ & 0.464 & 0.470 & $\mathbf{0.821}$ & 0.679\\

			$\textit{CRF40}$ & 0.493 & 0.510 & $\mathbf{0.826}$ & 0.688\\
			
			$\textit{GN(var=0.001)}$ & 0.572 & 0.548 & $\mathbf{0.827}$ & 0.682\\

			$\textit{GN(var=0.01)}$ & 0.533 & 0.510 & $\mathbf{0.698}$ & 0.649\\
			
			$\textit{GS(std=1.4)}$ & 0.515 & 0.507 & $\mathbf{0.826}$ & 0.683\\
			
			$\textit{GS(std=2.3)}$ & 0.514 & 0.507 & $\mathbf{0.768}$ & 0.666\\
			
			\hline
		\end{tabular}
    \caption{Comparison of the Laundering Cross-Domain Data. $\textit{ACC}$}
    \label{tab:cross-domain}
    \end{table}
    
    \begin{figure} 
	\centering  
	\subfigure{
		\label{frames}
		\includegraphics[width=0.45\textwidth]{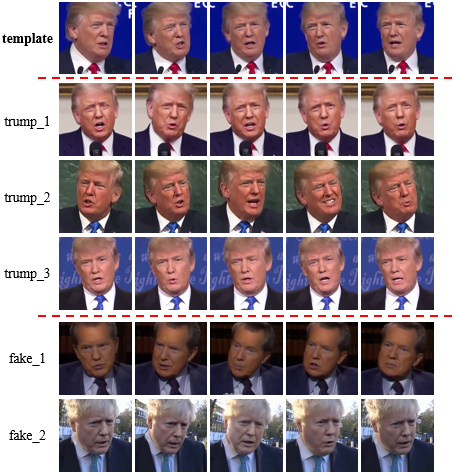}}

	\subfigure{
		\label{scatter}
		\includegraphics[width=0.45\textwidth]{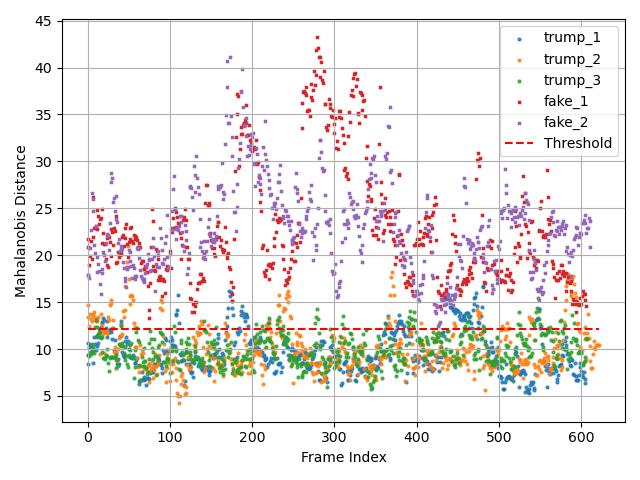}}
		
	\caption{Shown above are faces from genuine and face-swap videos of Donald Trump in the wild. Shown below is distribution of mahalanobis distances between the template and all faces in five videos.}
	\label{fig:wild}
    \end{figure}
    
\subsection{Face-Swap Detection in the Wild}
    Here, we further examine the effectiveness of our method in the wild. We gather four video clips of Donald Trump in various surroundings, including a template video. Moreover, two face-swap video clips replacing the faces of others with Trump are downloaded from the Internet. Fig~\ref{fig:wild} shows the faces from collected videos. We further extract 3D facial shape features of all frames in each video, and determine $\textit{Mahalanobis Distance}$ with the template.
    
    We utilize a scatter plot to show the experimental results of face-swap detection in the wild. According to Fig~\ref{fig:wild}, these points calculated from genuine faces are mainly distributed below the red line of the threshold. And they are effectively distinguished from those computed by the manipulated faces. Therefore, it demonstrates the effectiveness of our method in the wild. Moreover, note that the distribution of the points calculated from genuine faces does not alter with changes in the surroundings significantly. Consequently, the robustness of the feature expression of the facial shape is also demonstrated by the experiments.
    
\subsection{Experiment Results and Analysis}
    The experimental results on laundering data and laundering cross-domain data are presented in Table~\ref{tab:laundering} and Table~\ref{tab:cross-domain}. According to Table~\ref{tab:laundering}, our proposed approach shows strong robustness against laundering attacks. We train our model, $\textit{XceptionNet}$ and $\textit{EfficientNet}$ on $\textit{CRF23}$ data, and test them against various laundering counter-measures. Table~\ref{tab:laundering} summarizes the performance of all methods. Our proposed approach maintains similar performance when detecting different laundering data, while the effect of the pixel-level artifacts based methods are weakened. Moreover, our approach outperforms them on all laundering data and the accuracy of face-swap detection only has light fluctuation. 
    
    We further validate our method on cross-domain data generated by a previously unseen manipulation, $\textit{Deepfakes}$. In Table~\ref{tab:cross-domain}, our proposed method significantly outperforms the pixel-level artifacts based methods. $\textit{XceptionNet}$ and $\textit{EfficientNet}$ lose effects on the cross-domain data, while our proposed method shows comparable performance with the previous experiments, even on laundering cross-domain data. The performance of our method against different laundering counter-measures is similar. It further demonstrates the feature expression of 3D facial shape is robust on laundering and cross-domain data. 
    
\section{Conclusions} \label{conclusion}
     We tackle the task of face-swap detection with good robustness on laundering and cross-domain data. Our proposed method leverages the inconsistency between 3D facial shape information and facial appearance information to detect face-swap images. We further propose to utilize $\textit{Mahalanobis Distance}$ for measuring the inconsistency and present its superiority in our task. Moreover, we demonstrate that our approach is less vulnerable to laundering counter-measures and has good robustness against unseen face-swap methods. Finally, our proposed method also shows remarkable performance on the genuine and face-swap videos in the wild. In our method, 3D facial shape information plays a crucial role to detect face-swap images. In the future work, we will focus on more advanced facial shape estimation methods for better detection performance.

{\small
\bibliographystyle{ieee}
\bibliography{main}
}

\end{document}